\pdfoutput=1

\documentclass[11pt]{article}

\usepackage[]{ACL2023}

\usepackage{times}
\usepackage{latexsym}

\usepackage[T1]{fontenc}

\usepackage[utf8]{inputenc}

\usepackage{microtype}
\usepackage{inconsolata}
\usepackage{amsmath}
\usepackage{booktabs}
\usepackage{diagbox}
\usepackage{stfloats}
\usepackage[ruled]{algorithm2e}
\usepackage{graphicx}
\usepackage{color, soul}
\usepackage{float}
\usepackage{comment}
\title{Pre-trained Language Model with Prompts for Temporal Knowledge Graph Completion}
\newcommand*{\affaddr}[1]{#1} 
\newcommand*{\affmark}[1][*]{\textsuperscript{#1}}
\newcommand*{\email}[1]{\texttt{#1}}

\author{
Wenjie Xu\affmark[1], Ben Liu\affmark[1], Miao Peng\affmark[1], Xu Jia\affmark[1], Min Peng\affmark[1]\thanks{*Corresponding author}\\
\affaddr{\affmark[1]School of Computer Science, Wuhan University, China}\\
\email{\{vingerxu,liuben123,pengmiao,jia\_xu,pengm\}@whu.edu.cn}\\
}

\begin{document}
\maketitle
\begin{abstract}
Temporal Knowledge graph completion (TKGC) is a crucial task that involves reasoning at known timestamps to complete the missing part of facts and has attracted more and more attention in recent years. Most existing methods focus on learning representations based on graph neural networks while inaccurately extracting information from timestamps and insufficiently utilizing the implied information in relations. To address these problems, we propose a novel TKGC model, namely \textbf{P}re-trained Language Model with \textbf{P}rompts for \textbf{T}KGC (PPT). We convert a series of sampled quadruples into pre-trained language model inputs and convert intervals between timestamps into different prompts to make coherent sentences with implicit semantic information. We train our model with a masking strategy to convert TKGC task into a masked token prediction task, which can leverage the semantic information in pre-trained language models. Experiments on three benchmark datasets and extensive analysis demonstrate that our model has great competitiveness compared to other models with four metrics. Our model can effectively incorporate information from temporal knowledge graphs into the language models. The code of PPT is available at \url{https://github.com/JaySaligia/PPT}.

\end{abstract}

\section{Introduction}
In recent years, temporal knowledge graphs(TKGs) have attracted much attention. TKGs describe each fact in quadruple (\textit{subject, relation, object, timestamp}). Compared to static knowledge graphs, TKGs need to consider the impact of timestamps on events. For example,  (\textit{Donald Trump, PresidentOf, America, 2018}) holds while (\textit{Donald Trump, PresidentOf, America, 2022}) does not. 
\begin{figure}
    \centering
    \includegraphics[width=7cm, height=6cm]{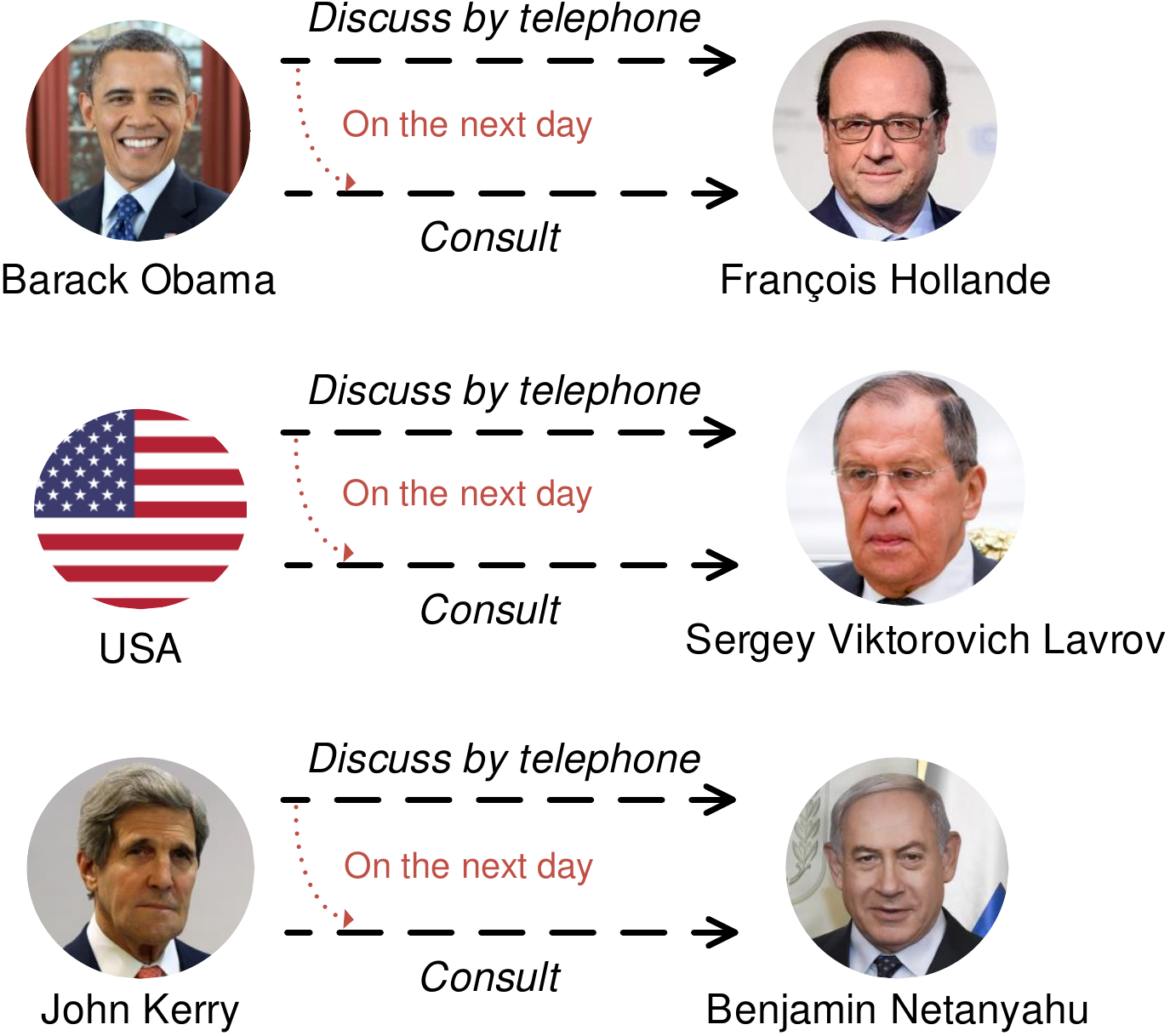}
    \caption{An example of the time-related semantic information between relations in three pairs of entities.}
    \label{fig:quadruples}
\end{figure}
There are missing entities or relations in the TKGs, therefore, temporal knowledge graph completion (TKGC) is one of the most important tasks of temporal knowledge graphs. TKGC task can be divided into two categories: interpolation setting and extrapolation setting\cite{RENET}. Interpolation setting aims to predict missing facts in the known timestamps while extrapolation setting attempts to infer future facts in the unknown ones. The latter is much more challenging, and in this work, we focus on the extrapolation setting. Some TKGC methods are developed from static knowledge graph completion (KGC). Such as adding time-aware score functions to KGC models\cite{TTransE,HyTE}, adding time-aware relational encoders to graph neural networks \cite{RENET, HIP}, adding a new time dimension to the tensor decomposition\cite{TNTComplEx,Tucker}, etc. In addition to those KGC-based models, reinforcement learning\cite{TITer}, time-aware neural network modeling\cite{Cygnet}, and other methods are also applied to TKGC. However, the methods mentioned above have some drawbacks, as follows:
(1) \textbf{Insufficient temporal information extraction from timestamps}. Most existing TKGC methods model timestamps explicitly or implicitly. Explicit modeling utilizes low-dimensional vectors to represent timestamps. However, real-life timestamps are infinite, and explicit modeling cannot learn all timestamp representations and predict events with unseen timestamps. Implicit modeling does not represent timestamps directly but takes timestamps to connect multiple knowledge graphs by determining the sequential relationship of these knowledge graphs. This approach often requires modeling the knowledge graph one by one, requires a lot of computation, and timestamps are used only to determine before and after things happen. All the above methods do not give full play to the temporal information of timestamps.  (2) \textbf{Insufficient information mining of associations in relations in TKGC}. Existing methods often focus on the structural information of the triples or quadruples when modeling KGs without enough consideration of the implied information in relations. This problem is particularly evident in TKGs because some relations contain information with potential temporal hints. As shown in Figure \ref{fig:quadruples}, between three different pairs of subject and object entities, after establishing relation \textit{Discuss by telephone}, one day apart, they all establish relation \textit{Consult}. If relation \textit{Discuss by telephone} is established between the same pair of entities, there is a high probability that they will establish relation \textit{Consult} within a short period. Among the entity pairs in ICEWS14, there are 10,887 types of relation pairs, out of which 2,652 exhibit obvious temporal correlations, where one relation in the pair high probably occurred before the other, and they have a stable time interval between them.

To address these problems, we propose a novel temporal knowledge graph completion method based on pre-trained language models (PLMs) and prompts. TKGs contain timestamps, and events occurring at different occurrence times have sequential relationships with each other, which are well-suited as inputs to sequence models. Inspired by the successful application of pre-trained language models in static knowledge graph representation\cite{KG-BERT, muilt, LAMA, PKGC}, we apply PLMs to temporal knowledge graph completion to get implicit semantic information. However, simply splicing entities and relations in the input of PLMs generates incoherent sentences, resulting in the inability to use PLMs\cite{PKGC} fully. Therefore, We sample the quadruples in TKGs and construct prompts for each type of timestamps, which we call \textbf{time-prompts}. Then we train PLMs with a masking strategy. In this way, TKGC can be converted into a masked token prediction task. 

The contributions of our work can be summarized as follows:
\begin{itemize}
    \item To the best of our knowledge, we are the first to convert the temporal knowledge graph completion task into the pre-trained language model masked token prediction task.
    \item We construct prompts for each type of interval between timestamps to better extract semantic information from timestamps. 
    \item We apply our experiments on a series of datasets of ICEWS and achieve satisfactory results compared to graph neural network learning methods. 
\end{itemize}

\section{Related Work}
\subsection{Static KG representation}
Static KG representation learning can roughly be divided into distance-based models, semantic matching models, graph neural network models, and PLM-based models. 

Distance-based models represent the relation of two entities into a translation vector, such as TransE\cite{TransE}, RotatE\cite{RotatE}, TransH\cite{TransH}. Semantic matching models measure the plausibility of facts using a triangular norm, such as RESCAL\cite{RESCAL}, Distmult\cite{Distmult}, ConvE\cite{ConvE}, ComplEx\cite{ComplEx}. Graph neural network models use feed-forward or convolutional layers or extend Laplacian matrix to learn the representation of entities and relations, such as GCN\cite{GCN}, GAT\cite{GAT}, R-GCN\cite{R-GCN}, SAGE\cite{SAGE}. 

PLM-based models have also been considered for static KG representation in recent years due to the ability to capture context information. KG-BERT\cite{KG-BERT} first introduces PLMs into static KG representation. Among PLM-based models, prompt-learning has attracted much attention in recent years and has been shown to be effective on many NLP tasks. LAMA\cite{LAMA} first introduces prompt-based knowledge to PLM. Other prompt-based models based on LAMA are dedicated to improving the presentation of KGs by automatic prompt generation or by adding soft prompts\cite{Prompt-1,Prompt-2,Prompt-3}. PKGC\cite{PKGC} proposes a new prompt-learning method to accommodate the open-world assumption based on KG-BERT. 

\subsection{Temporal KG representation}
Temporal KG representation requires consideration of how the facts are modeled in time series. Some temporal KG representation models are extended from static models. TTransE\cite{TTransE} incorporates temporal information into the scoring function based on TransE\cite{TransE}, and HyTE\cite{HyTE} extends TransH\cite{TransH} similarly. TNTComplEx\cite{TNTComplEx} extends ComplEx\cite{ComplEx} inspired by the CP decomposition of order-4 tensor. 

These expanded approaches consider timestamps as an additional dimension but lack consideration from a temporal perspective. Some models attempt to combine message-passing and temporal information to solve the problem. RE-NET\cite{RENET} applies R-GCN\cite{R-GCN} for message passing for each snapshot and then uses temporal aggregation across multiple snapshots. HIP Network\cite{HIP} utilizes structural information passing and temporal information passing to model snapshots. RE-GCN\cite{RE-GCN} uniformly encodes the evolutional representations representation of entities and relations corresponding to different timestamps to apply to the extrapolational TKGC task.

Besides, some models use other strategies to model TKG. CyGNet\cite{Cygnet} is divided into a copy mode and a generative mode to predict missing entities using neural networks with a time dictionary. TITer\cite{TITer} introduces reinforcement learning in TKG representation learning.
\section{Preliminary}
\textbf{Temporal Knowledge Graph} $\mathcal{G}$ is a set of networks of entities and relations that contain timestamps. It can be defined as $\mathcal{G}=\{\mathcal{E}, \mathcal{R}, \mathcal{T}, \mathcal{Q}\}$, where $\mathcal{E}$ is the set of entities, $\mathcal{R}$ is the set of relations and $\mathcal{T}$ is the set of timestamps. $\mathcal{Q} = \{(s,r,o,t)\}\subseteq \mathcal{E}\times \mathcal{R} \times \mathcal{E} \times \mathcal{T}$ is the quadruple set, where $s$ and $o$ are the subject entity (head entity) and object entity (tail entity), $r$ is the relation between them at timestamp $t$. $\mathcal{G}_t=\{(s,r,o)\subseteq \mathcal{E}\times \mathcal{R} \times \mathcal{E}\}$ is called the TKG snapshot at $t$, and it can be taken as a static KG filtering the triple set from $\mathcal{G}$ at $t$.

\textbf{Temporal Knowledge Graph completion} (TKGC) is the task of predicting the evolution of future KGs given KGs of a known period. Given a quadruple $(\mathrm{s}, \mathrm{r}, ?, t_n)$ or $(?,\mathrm{r},\mathrm{o},t_n)$, we have a set of known facts from TKG snapshots $G_{(t_i<t_n)}$ to predict the missing object entity or subject entity in the quadruple. The probability of prediction of missing the entity $o$ in quadruple $(\mathrm{s}, \mathrm{r}, ?, t_n)$ can be formalized as follows:
\begin{equation}
\label{raw}
    p(o\vert \mathcal{G}_{<t_n}, \mathrm{s},\mathrm{r},t_n).
\end{equation}
\section{Methodology}
\begin{figure*}[ht]
    \centering
    \includegraphics[width=15cm, height=9.5cm]{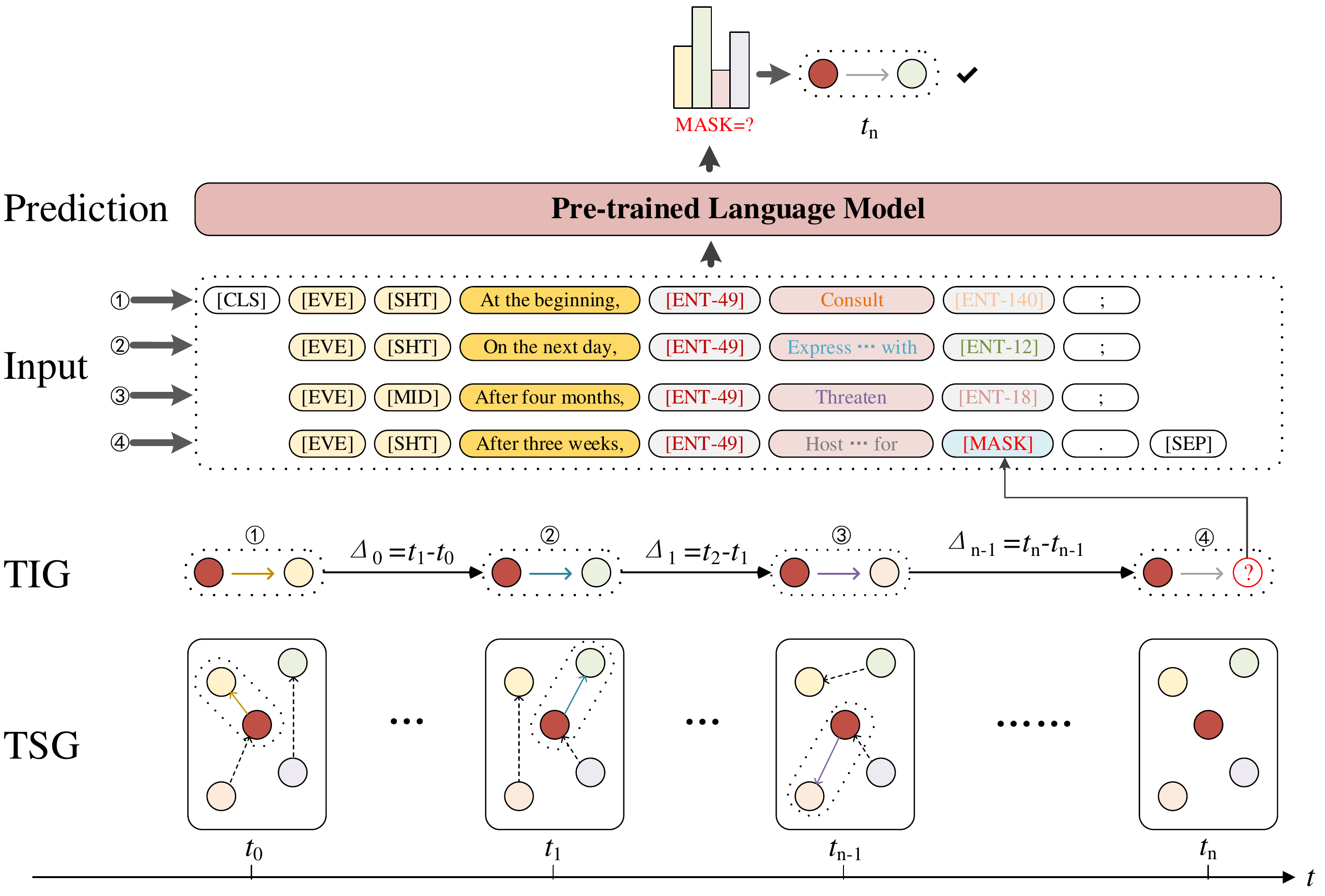}
    \caption{Illustration of PPT for TKGC. Quadruples are sampled and normalized to convert into PLM inputs with prompts. We calculate the time interval of adjacent quadruples in TSG to get TIG. We use the prompts to convert TIG into the input of PLM and then make the prediction for the mask. This way, The TKGC task is converted into a pre-trained language model masked token prediction task.}
    \label{fig:model}
\end{figure*}
In this paper, we propose PPT, a novel PLM-based model with prompts to solve TKGC task. The framework of our model is illustrated in Figure \ref{fig:model}. We sample quadruples and convert them into pre-trained language model inputs. The prediction of \textbf{[MASK]} token is the completed result.

\subsection{Prompts}
\label{sec:prompts}
\begin{figure}
    \centering
    \includegraphics[width=7cm, height=8cm]{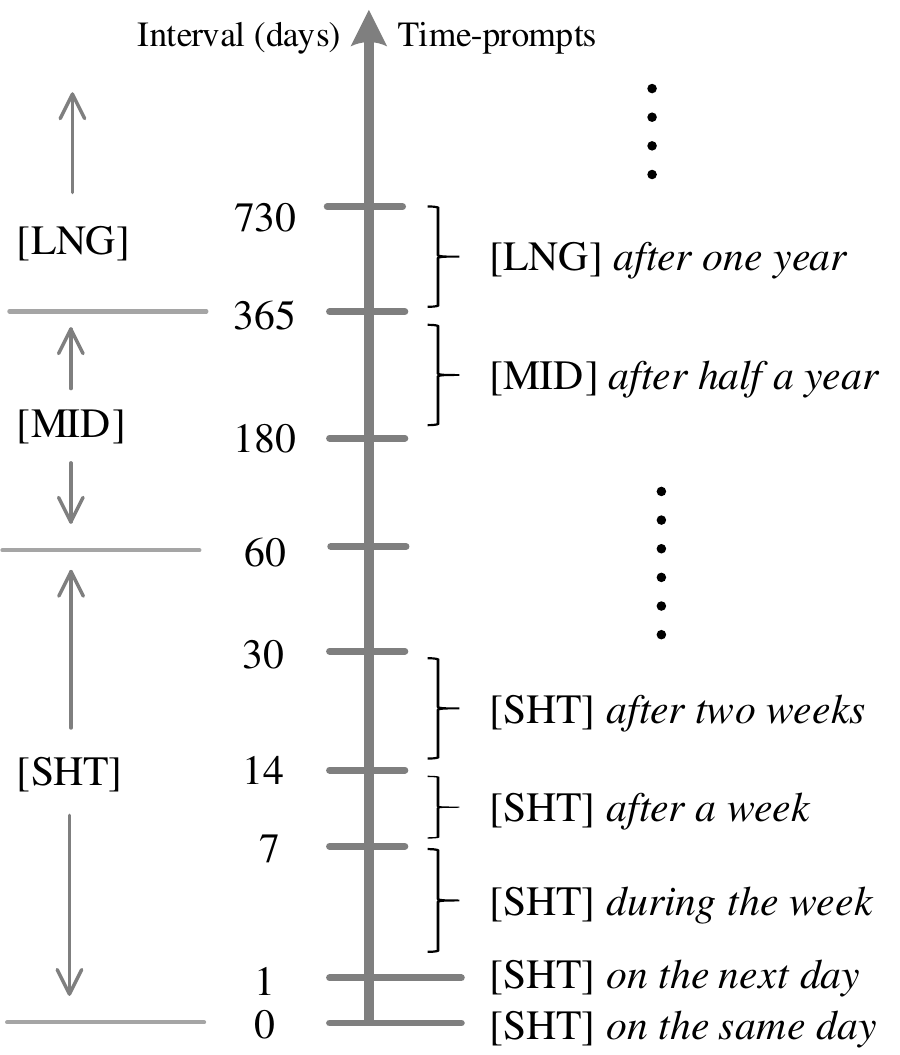}
    \caption{Illustration of \textit{interval-dictionary}. The left side of the vertical axis indicates the interval between two timestamps, and the right side indicates the time-prompts corresponding to the timestamp interval. [SHT] for short intervals ($\Delta_t \leq 60$), [MID] for medium intervals ($60 < \Delta_t \leq 365$), [LNG] for long intervals ($\Delta_t > 365$).}
    \label{fig:dictionary}
\end{figure}
We design different prompts for entities (ent-prompts), relations (rel-prompts), and timestamps (time-prompts) to convert quadruples into a form suitable for input to PLMs. We add a soft prompt \textbf{[EVE]} before the beginning of each fact tuple due to introducing soft prompts in the input sentences can improve the expressiveness of the sentences\cite{soft}.\\
\textbf{Ent-prompts}. We convert each entity into a special token \textbf{[ENT-i]} according to its index. We use a special token instead of the name of an entity because, in the prediction task, we need to predict the whole entity but not a part of it. To maintain the semantic information from entities, we do average pooling of embedding for all words in each entity as the initial embedding of its token.\\
\textbf{Rel-prompts}. For each relation, we convert it into its original phrase. It is worth noting that to maintain the coherence of sentences, we supplemented each relation with the preposition it was missing. For example, we supplement the relation \textit{Make a visit} to \textit{Make a visit to}.\\
\textbf{Time-prompts}. We convert the time interval between two timestamps into a phrase that can describe the period. We construct a dictionary called \textit{interval-dictionary}, which maps each period to a prompt. As shown in Figure \ref{fig:dictionary}, we convert each timestamp interval into a prompt. Each prompt contains two parts. 
The front part is a soft prompt indicating the length of time, such as \textbf{[SHT]} for a short time (less than 60 days), \textbf{[MID]} for a medium time (from 60 days to 365 days), and \textbf{[LNG]} for a long time (above 365 days); the back part is a statement describing the interval. During our analysis, we observed that news reports frequently use distinctive time descriptors to indicate time intervals, which inspired us to develop these prompts.
\subsection{Construction for Graphs}
Unlike sampling one fact tuple as input to a pre-trained language model in some static knowledge graph models\cite{KG-BERT, PKGC}, we sample multiple fact tuples simultaneously because we need to model the temporal relationship between facts. We take the head/tail entity for each quadruple in the training dataset and randomly sample each quadruple from the entire training dataset while fixing the head/tail entity. The sampled quadruples are then arranged in chronological order. We demonstrate different sampling strategies in \ref{subsec:0}. The sampled list is called \textit{Temporal Specialization Graph} (TSG). TSG can be described as a time-ordered sequence $TSG = [q_0,q_1 \dots, q_n],q_i=(s_i,r_i,o_i,t_i)\in \mathcal{Q}, t_i \leq t_{i+1}$. We have a total of three types of TSG, which are $TSG^s_{obj}$, $TSG^r_{sub}$ and $TSG^o_{rel}$:
\begin{equation}
    \begin{aligned}
       TSG^s_{obj}(n) =& [q_0^s,q_1^s \dots, q_n^s],\\
       q_i^s =& (obj,r_i,o_i,t_i)\in \mathcal{Q}, t_i \leq t_{i+1},\\
       TSG^r_{rel}(n) =& [q_0^r,q_1^r \dots, q_n^r],\\
       q_i^r =& (s_i,rel,o_i,t_i)\in \mathcal{Q}, t_i \leq t_{i+1},\\
       TSG^o_{sub}(n) =& [q_0^o,q_1^o \dots, q_n^o],\\
       q_i^o= & (s_i,r_i,sub,t_i)\in \mathcal{Q}, t_i \leq t_{i+1},
    \end{aligned}
\end{equation}
where we fix object entity $obj$ to sample $TSG^s_{obj}$, fix subject entity $sub$ to get sample $TSG^r_{sub}$, and fix relation $rel$ to sample $TSG^o_{rel}$. We set a minimum sampling quadruple number $\mathrm{K}_{min}$ and a maximum sampling quadruple number $\mathrm{K}_{max}$. 

The timestamps in TSGs are independent and cannot reflect the time relationship between events. We convert each TSG to a \textit{Time Interval Graph} (TIG) by calculating the time interval of adjacent quadruples. We take the earliest time in TSG as the initial time $\tau_0$ and calculate the time interval between the timestamp in $(s_i,r_i,o_i,t_i)$ and the timestamp in $(s_{i-1},r_{i-1},o_{i-1},t_{i-1})$ as the new timestamp $\tau_{i}$:
\begin{equation}
\begin{array}{l}
TIG_{*, *=\{s,r,o\}} = [p_0^*,p_1^*, \dots, p_n^*], \\
p_i^* = (q_i^*(s,r,o), \tau_i), \\
\left\{\begin{array}{l}
\tau_o = 0 \\
\tau_i = t_{i} - t_{i-1}
\end{array},\right.
\end{array}
\end{equation}
where $q_i^*(s,r,o)$ means keeping the fact triple $(s_i,r_i,o_i)$ of $q_i^*$.

\subsection{Training}
The algorithm of our training strategy can be summarized in Algorithm \ref{algorithm}. We do not train each quadruple separately in the training set for each epoch because we believe that independent quadruples cannot provide temporal information in TKGs. We sample each entity multiple times by fixing it at the object entity position and the subject entity position, thus generating TSGs of entities. Similarly, we fix the relations in the quadruples and, for each relation generate the TSGs of the relations. Then we convert all the TSGs to TIGs. For each quadruple in a TIG, we convert the entities, relation, and time interval into PLM inputs with prompts described in Section.\ref{sec:prompts}. We use a pre-trained language model with the masking strategy (also known as a masked language model, MLM)\cite{BERT} to train our model. Masked language models aim to predict masked parts based on their surrounding context. When training, we mask 30\% of tokens in an input sequence.

\begin{algorithm}
\scriptsize
\caption{Training for PPT}
\KwIn{TKG $\mathcal{G}$ with training data, maximum number of epochs $max\_epochs$, maximum number of sampling TSG of one entity or one relation $\mathrm{B}$,  minimum sampling sequence length $\mathrm{K}_{min}$, maximum sampling sequence length $\mathrm{K}_{max}$.}
\Repeat{
$epoch = max\_epochs$\;}{
$epoch \leftarrow 1$\;
$\mathcal{S} = \{\}$\;
\For{$b \leftarrow 1$ to $\mathrm{B}$}{
    \ForEach{${ent} \in\mathcal{E}$}{
        \tcp{sample TSG for entities}
        $k=random(\mathrm{K}_{min},\mathrm{K}_{max})$\;
        Sample a $TSG_{ent}$ with length = k\;
        Convert $TSG_{ent}$ into $TIG_{ent}$\;
        add $TIG_{ent}$ to $\mathcal{S}$\;
        }
    \ForEach{${rel} \in\mathcal{R}$}{
        \tcp{sample TIG for relations}
        $k=random(\mathrm{K}_{min},\mathrm{K}_{max})$\;
        Sample a $TSG_{rel}$ with length = k\;
        Convert $TSG_{rel}$ into $TIG_{rel}$\;
        add $TIG_{rel}$ to $\mathcal{S}$\;
    }
}

\ForEach{$TIG \in \mathcal{S}$}{
    \tcp{convert TIG into input with prompts}
    $seq = \mathbf{Prompt}(TIG)$\;
    \tcp{train in PLM with masking strategy}
    $\mathbf{MASK\_TRAIN}(PLM(seq))$\;
}
$epoch \leftarrow epoch + 1$\;
}
\label{algorithm}
\end{algorithm}

\subsection{Objective optimization discussion}
The distribution of all facts in Eq \ref{raw} can be considered as the joint distribution of facts on all timestamps:
\begin{equation}
\label{pG}
\begin{aligned}
   p(\mathcal{G}_{<t_n}) =& p(\mathcal{G}_{t_0}, \mathcal{G}_{t_1},\cdots, \mathcal{G}_{t_{n-1}})\\
   =& \prod_t \prod_{(s_t, r_t, o_t) \in \mathcal{G}_t} p\left(s_t, r_t, o_t \mid G_{<t_n}\right).
\end{aligned}
\end{equation}

It is not realistic to focus on all quadruples in the TKG. When predicting the missing subject entities, we fix the object entities because relations in the neighborhood are of most interest to entities. Further, we simulate the original quadruple distribution by sampling, thus Eq \ref{pG} can be approximated as:
\begin{equation}
\begin{aligned}
    p(\mathcal{G}_{<t_n}) \approx & \prod_t \prod_{(\mathrm{s}, r_t, o_t) \in \mathcal{G}_t} p\left(\mathrm{s}, r_t, o_t \mid G_{<t_n}\right)\\
    \approx & \prod_{k=1}^K p\left(\mathrm{s}, r_k, o_k \mid G_{<t_n}\right)\\
    \approx & \prod_{k=1}^K p\left(TSG_{\mathrm{s}}^s[k] \mid G_{<t_n}\right)\\
    \approx & \prod_{k=1}^K p\left(TIG_{\mathrm{s}}^s[k] \mid G_{<t_n}\right),
\end{aligned}
\end{equation}
where $K$ is the number of sampling.

We calculate the generation probability of the quadruples by the pre-trained language model's ability to predict unknown words. We use $seq_k$ to present the converted inputs with prompts of $TIG_{\mathrm{s}}^s[k]$:
\begin{equation}
\label{inputs}
seq = \mathbf{Prompt}(TIG_{\mathrm{s}}^s[k]).
\end{equation}

For example, as illustrated in Figure \ref{fig:model}, here are two quadruples in TSG:$(49,62,12,2)$ in timestamp $t_1$ and $(49,38,18,130)$ in timestamp $t_{n-1}$, the time interval between them is 128 days, $\Delta_1 = t_{n-1}-t_1$. Then the quadruple $(49,38,18,128)$ in TIG can be converted into an input sentence with prompts: \textbf{[EVE]} \textbf{[MID]} \textit{After four months}, \textbf{[ENT-49]} \textit{Threaten} \textbf{[ENT-18]}. 

The formalization of prediction can be defined as follows:
\begin{equation}
\label{MLM}
\begin{aligned}
    & \prod_{k=1}^K p\left(TIG_{\mathrm{s}}^s[k] \mid G_{<t_n}\right) \\
    =& \prod_{k=1}^K p(PLM(seq_k)),
\end{aligned}
\end{equation}
where $PLM(\cdot)$ means inputting a sequence into the pre-trained language model.

Combining Eq \ref{raw} and Eq \ref{MLM}, we convert the TKGC task into an MLM prediction task:
\begin{equation}
\label{all}
\begin{aligned}
    & p(o\vert \mathcal{G}_{<t_n}, \mathrm{s},\mathrm{r},t_n) \\
    \approx & \prod_{k=1}^K p(PLM(seq_k)) \\
    \cdot & p(PLM(\mathbf{Prompt}(\mathrm{s}, \mathrm{r}, t_n))),
\end{aligned}
\end{equation}
where $\mathbf{Prompt}(\cdot)$ means converting entities, relations, and timestamps into input sequences for PLM. 

By Eq \ref{all}, the original knowledge-completion task can be equated to the pre-trained language model masked token prediction task.

\section{Experiments}
\begin{table*}[htbp]
\centering
\begin{tabular}{@{}lllllll@{}}
\toprule
\textbf{Dataset} & $\mathcal{E}$ & $\mathcal{R}$ & \textbf{\#Granularity} & \textbf{\#Train} & \textbf{\#Valid} & \textbf{\#Test} \\ \midrule
ICEWS05-15       & 10094         & 251           & 24 (hours)               & 368868           & 46302            & 46159           \\
ICEWS14          & 6869         & 230           & 24 (hours)               & 74845            & 8514            & 7371           \\
ICEWS18          & 23033    & 256           & 24 (hours)               & 373018           & 45995            & 49545           \\ \bottomrule
\end{tabular}
\caption{Statistics of the datasets we use.}
\label{tab:datasets}
\end{table*}
\subsection{Experimental Setup}
\begin{table}[ht]
\scalebox{0.9}{
\begin{tabular}{@{}llll@{}}
\toprule
dataset    & seq\_len & min\_sample & max\_sample \\ \midrule
ICEWS05-15 & 256      & 2           & 16          \\
ICEWS14    & 256      & 2           & 12          \\
ICEWS18    & 256      & 2           & 16          \\ \bottomrule
\end{tabular}}
\caption{Parameters for datasets.}
\label{tab:set}
\end{table}
\textbf{Datasets.} Intergrated Crisis Early Warning System (ICEWS)\cite{ICEWS} is a repository that contains coded interactions between socio-political actors with timestamps. We utilize three TKG datasets based on ICEWS named ICEWS05-15(\cite{TA-DistMult}; from 2005 to 2015), ICEWS14(\cite{TA-DistMult}; from 1/1/2014 to 12/31/2014) and ICEWS18(\cite{ICEWS}; from 1/1/2018 to 10/31/2018) to perform evaluation. Statistics of these datasets are listed in Table \ref{tab:datasets}. \\ \textbf{Evaluation Protocals.} Following prior work\cite{RE-GCN}, we split each dataset into a training set, validation set, and testing set in chronological order following extrapolation setting. Thus, we guarantee that \textit{timestamps of train} < \textit{timestamps of valid} < \textit{timestamps of test}. Some methods\cite{RENET, Cygnet, TeMP} apply \textit{filter schema} to evaluate the results by removing all the valid facts that appear in the training, validation, or test sets from the ranking list. Since TKGs are evolving in time, the same event can occur at different times\cite{RE-GCN}. Therefore, we apply \textit{raw schema} to evaluate our experiments by removing nothing. We report the result of Mean Reciprocal Ranks(MRR) and Hits@1/3/10 (the proportion of correct test cases that are ranked within the top 1/3/10) of our approach and baselines following \textit{raw schema}.\\ \textbf{Baselines.} We compare our model with two categories of models: static KGC models and TKGC models. We select DistMult\cite{Distmult}, ComplEx\cite{ComplEx}, R-GCN\cite{R-GCN}, ConvE\cite{ConvE}, ConvTransE\cite{ConvTransE}, RotatE\cite{RotatE} as static models. We select HyTE\cite{HyTE}, TTransE\cite{TTransE}, TA-DistMult\cite{TA-DistMult}, RGCRN\cite{RGCRN}, CyGNet\cite{Cygnet}, RE-NET\cite{RENET}, RE-GCN\cite{RE-GCN} as baselines of TKGC. \\
\textbf{Hyperparameters.} We use bert-base-cased\footnote{\url{https://huggingface.co/bert-base-cased}} as our pre-trained model. Bert-base-cased has been pre-trained on a large corpus of English data in a self-supervised fashion. Bert-base-cased has a parameter size of 110M with 12 layers and 16 attention heads, and its hidden embedding size is 768. Without loss of generality, we also list other pre-trained models in \ref{subsec:2}. The input sequence length, min sampling number, and max sampling number of each dataset are listed in Table \ref{tab:set}. When training, we mask 30\% tokens randomly, and we choose AdamW as our optimizer. The learning rate is set as 5e-5. We make a detailed analysis of the parameters in \ref{subsec:1}.
\begin{table*}[ht]
\centering
\scalebox{0.75}{
\begin{tabular}{@{}lllllllllllll@{}}
\toprule
 &
  \multicolumn{4}{l}{ICEWS05-15} &
  \multicolumn{4}{l}{ICEWS14} &
  \multicolumn{4}{l}{ICEWS18} \\ \cmidrule(l){2-13} 
Method &
  {\color[HTML]{000000} MRR} &
  {\color[HTML]{000000} Hits@1} &
  {\color[HTML]{000000} Hits@3} &
  {\color[HTML]{000000} Hits@10} &
  {\color[HTML]{000000} MRR} &
  {\color[HTML]{000000} Hits@1} &
  {\color[HTML]{000000} Hits@3} &
  {\color[HTML]{000000} Hits@10} &
  {\color[HTML]{000000} MRR} &
  {\color[HTML]{000000} Hits@1} &
  {\color[HTML]{000000} Hits@3} &
  {\color[HTML]{000000} Hits@10} \\ \midrule
DistMult &
  19.91 &
  5.63 &
  27.22 &
  47.33 &
  20.32 &
  6.13 &
  27.59 &
  46.61 &
  13.86 &
  5.61 &
  15.22 &
  31.26 \\
ComplEx &
  20.26 &
  6.66 &
  26.43 &
  47.31 &
  22.61 &
  9.88 &
  28.93 &
  47.57 &
  15.45 &
  8.04 &
  17.19 &
  30.73 \\
R-GCN &
  27.13 &
  18.83 &
  30.41 &
  43.16 &
  28.03 &
  19.42 &
  31.95 &
  44.83 &
  15.05 &
  8.13 &
  16.49 &
  29.00 \\
ConvE &
  31.40 &
  21.56 &
  35.70 &
  50.96 &
  30.30 &
  21.30 &
  34.42 &
  47.89 &
  22.81 &
  13.63 &
  25.83 &
  41.43 \\
ConvTransE &
  30.28 &
  20.79 &
  33.80 &
  49.95 &
  31.50 &
  22.46 &
  34.98 &
  50.03 &
  23.22 &
  14.26 &
  26.13 &
  41.34 \\
RotatE &
  19.01 &
  10.42 &
  21.35 &
  36.92 &
  25.71 &
  16.41 &
  29.01 &
  45.16 &
  14.53 &
  6.47 &
  15.78 &
  31.86 \\ \midrule
HyTE &
  16.05 &
  6.53 &
  20.20 &
  34.72 &
  16.78 &
  2.13 &
  24.84 &
  43.94 &
  7.41 &
  3.10 &
  7.33 &
  16.01 \\
TTransE &
  16.53 &
  5.51 &
  20.77 &
  39.26 &
  12.86 &
  3.14 &
  15.72 &
  33.65 &
  8.44 &
  1.85 &
  8.95 &
  22.38 \\
TA-DistMult &
  27.51 &
  17.57 &
  31.46 &
  47.32 &
  26.22 &
  16.83 &
  29.72 &
  45.23 &
  16.42 &
  8.60 &
  18.13 &
  32.51 \\ 
RGCRN &
  35.93 &
  26.23 &
  40.02 &
  54.63 &
  33.31 &
  24.08 &
  36.55 &
  51.54 &
  23.46 &
  14.24 &
  26.62 &
  41.96 \\
CyGNet &
  35.46 &
  25.44 &
  40.20 &
  54.47 &
  35.45 &
  26.05 &
  39.91 &
  53.20 &
  26.46 &
  16.62 &
  30.57 &
  \ul{45.58} \\
RE-NET &
  36.86 &
  26.24 &
  41.85 &
  57.60 &
  35.77 &
  25.99 &
  40.10 &
  54.87 &
  26.17 &
  16.43 &
  29.89 &
  44.37 \\
RE-GCN &
  \ul {38.27} &
  \ul {27.43} &
  \ul {43.06} &
  \textbf{59.93} &
  \ul {37.78} &
  \ul {27.17} &
  \ul {42.50} &
  \textbf{58.84} &
  \textbf{27.51} &
  \textbf{17.82} &
  \textbf{31.17} &
  \textbf{46.55} \\ \midrule
PPT &
  \textbf{38.85} &
  \textbf{28.57} &
  \textbf{43.35} &
  \ul {58.63} &
  \textbf{38.42} &
  \textbf{28.94} &
  \textbf{42.5} &
  \ul {57.01} &
  \ul {26.63} &
  \ul {16.94} &
  \ul {30.64} &
   45.43 \\ \bottomrule 
\end{tabular}}
\caption{Results on three datasets. The best results are boldfaced, and the second best ones are underlined. The results of baselines are from RE-GCN\cite{RE-GCN}.}
\label{tab:result}
\end{table*}
\subsection{Results}
We report the results of PPT and baselines in Table \ref{tab:result}. 

It can be observed that PPT outperforms all static models much better. Compared with ConvTransE, which has the best results among static models, we achieve 28.3\%, 21.97\%, and 14.69\% improvement with MRR metric in the three datasets, respectively. We believe temporal information matters in TKGC tasks, while static models do not utilize temporal information. 

As can be seen that PPT performs better than HyTE, TTransE, and TA-DistMult. These models are under the interpolation setting. For instance, we achieve 41.22\%, 46.53\%, and 62.18\% improvements with MRR metric in the three datasets compared to TA-DistMult. We believe that HyTE and TA-DistMult only focus on independent graphs and do not establish the temporal correlation between graphs. TTransE embeds timestamps into the scoring function while not taking full advantage of them. 

With MRR, Hits@1, and Hits@3 metrics on ICEWS05-15 and ICEWS14, PPT achieves the best results compared to other TKGC models. For instance, PPT improves 6.5\% over the second-best result with Hit@1 metric. On ICEWS18, PPT has a slight gap with the best model RE-GCN. We believe this is because ICEWS18 has more entities than other datasets. GNN-based models using the message-passing mechanism have better learning ability for such graphs with many nodes. Furthermore, RE-GCN adds additional edges to assist learning for the static parts of the graph.

Besides the masking strategy for our model, we also attempt other forms of application for pre-trained language models, which are illustrated in \ref{subsec:2}. 
\subsection{Ablation study}
\begin{table*}[ht]
\centering
\scalebox{0.7}{
\begin{tabular}{@{}lllllllllllll@{}}
\toprule
Method & \multicolumn{4}{l}{ICEWS05-15} & \multicolumn{4}{l}{ICEWS14}   & \multicolumn{4}{l}{ICEWS18}   \\ \cmidrule(l){2-13} 
    & MRR   & Hits@1 & Hits@3 & Hits@10 & MRR   & Hits@1 & Hits@3 & Hits@10 & MRR   & Hits@1 & Hits@3 & Hits@10 \\ \midrule
PPT & \textbf{38.85} & \textbf{28.57}  & \textbf{43.35}  & \textbf{58.63}   & \textbf{38.42} & \textbf{28.94}  & \textbf{42.5}   & \ul{57.01}   & \textbf{26.63} & \textbf{16.94}  & \textbf{30.64}  & \textbf{45.43}   \\
PPT w/o prompts         & \ul{38.44}  & \ul{28.09} & \ul{43.09} & \ul{58.46} & \ul{38.24} & \ul{28.52} & \ul{42.4}  & \textbf{57.31} & \ul{25.44} & \ul{15.68} & \ul{29.26} & \ul{44.88} \\
PPT rand prompts        & 37.43  & 27.05 & 42.16 & 57.49 & 36.84 & 26.89 & 41.41 & 55.73 & 24.22 & 14.31 & 28.09 & 44.32 \\ \bottomrule
\end{tabular} }
\caption{Ablation experiments results of PPT. The best results are boldfaced and the second best ones are underlined.}
\label{tab:ablation}
\end{table*}
To investigate the contribution of time-prompts in our model, we conduct ablation studies for our model by testing all datasets under the same parameter settings of different variants. The experiment results are shown in Table \ref{tab:ablation}.

\textit{PPT w/o prompts} denotes PPT without time-prompts. In this variant, we set all timestamps as 0. To ensure that the sequence length does not affect the experiments, we replaced all the time-prompts with \textit{on the same day}.  \textit{PPT w/o prompts} gets worse results than raw PPT with all metrics on three datasets except with Hits@10 on ICEWS14. ICEWS14 has a smaller number of entities and data size than the other two datasets, so it is possible to achieve better results in some metrics after removing the timestamps.

\textit{PPT rand prompts} denotes PPT with random timestamps set. We replace raw timestamps in quadruples with other timestamps randomly. Random timestamps should not affect the results if our model does not learn the timestamp information correctly. As shown in Table \ref{tab:ablation}, the raw model shows better results than this variant on all metrics. 

These experiments demonstrate that applying time-prompts in our model can benefit the learning of temporal information between events.

\subsection{Analysis}
\subsubsection{Attention analysis}
\begin{figure}[ht]
    \centering
    \includegraphics[scale=0.55]{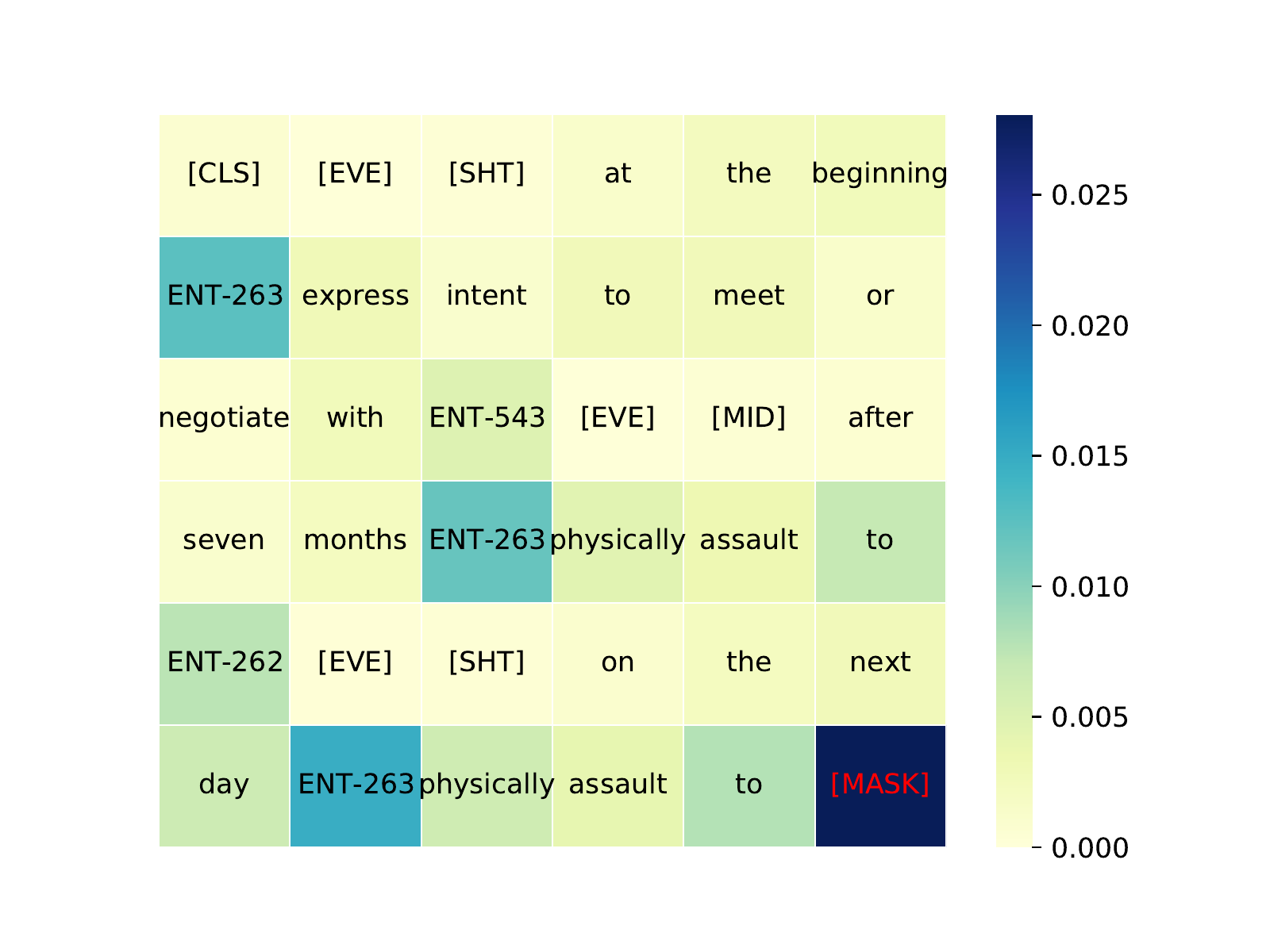}
    \caption{Illustrations of attention patterns of PPT. The quadruple that needs to be completed is $(263,104,?,7536)$, we sample 2 quadruples with earlier timestamps than the test example and fixed object entities. Transparencies of colors reflect the attention scores of other tokens to \textbf{[MASK]}.}
    \label{fig:attention}
\end{figure}
To visually show that our model can learn from temporal knowledge graphs, as shown in Figure \ref{fig:attention}, we visualize attention patterns of PPT. We need to complete the missing tail entity in a test quadruple $(263,104,?,7536)$. As mentioned, we sample data from earlier than timestamp 7536 to form the input sequence and obtain the attention weights from the pre-trained model. In this example, the ground truth is \textbf{[ENT-262]}. We observe that in our model, the prediction of \textbf{[MASK]} is made by considering all the previous sampling samples together. PPT notes that the same relation \textit{physical assault to} occurred a day earlier and captures the temporal information from token \textbf{the}, \textbf{next}, and \textbf{day}. Therefore, PPT can make correct predictions based on historical events and chronological relationships.
\subsubsection{Time-sensitive relation analysis}
Using ICEWS05-15 as an example, we analyze the time-sensitive relations present in the dataset. For different relations between the same pairs of entities, there is a clear order of occurrence among some of them. For example, the relation \textit{Obstruct passage, block} is always followed by ones related to assistance, such as \textit{Appeal for aid}, \textit{Appeal for humanitarian aid}, and \textit{Provide humanitarian aid}. Similarly, the relation \textit{Acknowledge or claim responsibility} is always followed by those related to negotiation, such as \textit{Express intent to cooperate militarily}, \textit{Meet at a 'third' location}, and \textit{Demand material cooperation}. We provide more examples in \ref{subsec:4}.

To verify the superiority of PPT in handling time-sensitive relations, a new test dataset named \textit{ICEWS05-filter} is constructed from ICEWS05-15. Specifically, we select relations that have a clear chronological order within a predefined time window, resulting in a total of 139 relations. Only the quadruples containing these selected relations are retained to construct the new dataset. As demonstrated in Table \ref{tab:sensitive}, PPT achieves better performance when evaluated on the constructed test dataset, indicating its advantage in handling time-sensitive relations.

\begin{table}[h]
\centering
\scalebox{0.8}{
\begin{tabular}{@{}lllll@{}}
\toprule
Dataset  & MRR   & Hits@1 & Hits@3 & Hits@10 \\ \midrule
ICEWS05-15 & 38.85 & 28.57  & 43.35  & 58.63   \\
ICEWS05-filter    & 39.4  & 29.02  & 43.91  & 59.31   \\ \bottomrule
\end{tabular}}
\caption{Results of PPT on ICEWS05-15 and ICEWS05-filter.}
\label{tab:sensitive}
\end{table}

\section{Conclusions}
This paper proposes a novel temporal knowledge graph completion model named pre-trained language model with prompts for TKGC (PPT). We use prompts to convert entities, relations, and timestamps into pre-trained model inputs and turn TKGC problem into a masked token prediction problem. This way, we can extract temporal information from timestamps accurately and sufficiently utilize implied information in relations. Our proposed method achieves promising results compared to other temporal graph representation learning methods on three benchmark TKG datasets. For future work, we plan to improve the sampling method in temporal knowledge graphs to get more time-specific inputs. We are also interested in combining GNNs and pre-trained language models in temporal knowledge graph representation learning.
\section*{Limitations}
This paper proposes a pre-trained language model with prompts for temporal knowledge graph completion. However, there are some limitations in our method: 1) Our prompts in the temporal knowledge graphs, especially the time-prompts, are built manually. It needs to be reconstructed manually for different knowledge graphs. We are exploring a way to build prompts in temporal knowledge graphs automatically. 2) Our model uses a random sampling method, which suffers from the problem of few high-quality training samples and high sample noise. For future work, a more effective way to sample is worth exploring.
\section*{Acknowledgements}
We would like to thank all the anonymous reviewers for their valuable and insightful comments. This work was supported by National Key Research and Development Program of China (No.2021ZD0113304), General Program of  Natural Science Foundation of China (NSFC) (Grant No.62072346), Key R\&D Project of Hubei Province (Grant NO.2021BBA099, NO.2021BAA029) and Application Foundation Frontier Project of Wuhan (Grant NO.2020010601012168). Our work was founded by Joint\&Laboratory on Credit Technology.
\section*{Ethics Statement}
All steps and data described in our paper follow the ACL Ethics Policy\footnote{\url{https://www.aclweb.org/portal/content/acl-code-ethics}}.
\bibliography{anthology,custom}
\bibliographystyle{acl_natbib}
\clearpage
\appendix
\section{Appendix}
\label{sec:appendix}
\subsection{Sampling Analysis}
\label{subsec:0}
We design two sampling strategies, one is the uniform sampling strategy, and the other is the frequency-based sampling strategy. The uniform sampling strategy assigns equal sampling weights to each entity. The frequency-based sampling strategy assigns different weights to each entity based on the different frequencies of each entity appearing in the dataset, where entities with higher occurrences have a higher probability of being sampled. As shown in Table \ref{tab:strategy}, the frequency-based sampling strategy has better results on ICEWS14. We believe this is because if an entity appears frequently, it is more likely to have relations with other entities and should get more attention.

\begin{table}[h]
\centering
\scalebox{0.75}{
\begin{tabular}{@{}lllll@{}}
\toprule
Strategy        & MRR   & Hits@1 & Hits@3 & Hits@10 \\ \midrule
uniform         & 34.87 & 25.37  & 38.77  & 53.33   \\
frequency-based & 38.42 & 28.94  & 42.5   & 57.01   \\ \bottomrule
\end{tabular}}
\caption{Results of different sampling strategies of PPT on ICEWS14.}
\label{tab:strategy}
\end{table}

\subsection{Hyperparameter Analysis}
\label{subsec:1}
To test the effect of different sequence lengths and the maximum number of samples on the effect of the model, we analyze these hyperparameters on ICEWS14. Due to GPU performance limitations, we do not perform experiments on longer sequences. 

As shown in Table \ref{tab:hyper}, we get the best results with setting $seq\_len=256,max\_sample=12$. We believe that the effect of sequence length is small while the number of samples matters. A larger number of samples can provide more semantic contextual information for the prediction but overly lengthy sampling can cause a decline in effectiveness by not focusing on the most effective information in learning. 

\begin{table}[h]
\centering
\scalebox{0.75}{
\begin{tabular}{@{}llllll@{}}
\toprule
seq\_len & max\_sample & MRR   & Hits@1 & Hits@3 & Hits@10 \\ \midrule
128      & 2           & 35.33 & 25.71  & 39.56  & 53.83   \\
128      & 4           & 37.21 & 27.59  & 41.08  & 56.3    \\
128      & 8           & 37.67 & \ul{28.16}  & 41.73  & 56.22   \\
256      & 8           & 37.67 & 27.78  & \ul{42.31}  & 56.72   \\
256      & 12          & \textbf{38.42} & \textbf{28.94}  & \textbf{42.5}   & \textbf{57.01}   \\
256      & 16          & \ul{37.72} & 27.74  & 42.1   & \ul{56.91}   \\ \bottomrule
\end{tabular}}
\caption{Results of different hyperparameters of PPT on ICEWS14. The best results are boldfaced and the second best ones are underlined.}
\label{tab:hyper}
\end{table}
\subsection{Variants}
\label{subsec:2}
In addition to the model we propose in the paper, we also try some variants, all experiments are done with $seq\_len=256,max\_sample=12$ on ICEWS14. As demonstrated in Table \ref{tab:variants}, PPT\_CLS does not use the mask training strategy but takes \textbf{[CLS]} to do classification with a fully connected layer as the decoder; PPT\_LSTM uses a bi-directional LSTM to encode all tokens, max-pool the out embeddings, and use a fully-connected layer as a decoder. These models do not get satisfactory results compared to our raw model. 

PPT\_CLS only uses sequence embedding to predict the result is not enough because the sequence embedding is suitable for classification task which needs to be focused on the whole input sequence. However, in our task, we need to consider the impact of each token. For PPT\_LSTM, we believe that the representation learned by the pre-trained language model is high-level semantic knowledge, especially when additional tokens (entities and relations) are added. Simple neural network models are unable to capture this high-level semantic knowledge and instead cause a decrease in effectiveness.

\begin{table}[h]
\centering
\scalebox{0.75}{
\begin{tabular}{@{}lllll@{}}
\toprule
Variants  & MRR   & Hits@1 & Hits@3 & Hits@10 \\ \midrule
PPT\_CLS  & 32.81 & 23.62  & 36.81  & 51.12   \\
PPT\_LSTM & 32.6  & 23.61  & 36.54  & 50.06   \\
PPT       & 38.42 & 28.94  & 42.5   & 57.01   \\ \bottomrule
\end{tabular}}
\caption{Variants of PPT.}
\label{tab:variants}
\end{table}
\subsection{Different PLMs}
\label{subsec:3}
Besides \textit{bert-base-cased}, we also attempt other pre-trained language models: bert-base-uncased\footnote{\url{https://huggingface.co/bert-base-uncased}} and bert-large-cased\footnote{\url{https://huggingface.co/bert-large-uncased}}. As shown in Table \ref{tab:PLMs}. All experiments are done with setting $seq\_len=128,min\_sample=2,max\_sample=8$ on ICEWS14. We find that the experimental results with different PLMs are similar, indicating that our approach does not rely on a specific pre-trained language model and has the ability to generalize.

\begin{table}[h]
\centering
\scalebox{0.75}{
\begin{tabular}{@{}lllll@{}}
\toprule
PLMs              & MRR   & Hits@1 & Hits@3 & Hits@10 \\ \midrule
bert-base-cased   & 37.67 & 28.16  & 41.73  & 56.22   \\
bert-base-uncased & 37.75 & 28.06  & 41.74  & 56.84   \\
bert-large-cased  & 37.36 & 27.39  & 41.39  & 57.59   \\ \bottomrule
\end{tabular}}
\caption{Experiments on different PLMs.}
\label{tab:PLMs}
\end{table}

\subsection{Pre-relations and post-relations}
\label{subsec:4}
For one pair of entities, if relation \textit{rel-A} always occurs before relation \textit{rel-B}, \textit{rel-A} is called a pre-relation and \textit{rel-B} is called a post-relation. Table \ref{tab:rels} shows some of these relations.

\begin{table*}[ht]
\centering
\begin{tabular}{@{}ll@{}}
\toprule
Pre-relation                                                 & Post-relation                                                              \\ \midrule
{\color[HTML]{333333} Demonstrate for policy change}         & {\color[HTML]{333333} fight with small arms and light weapons}             \\
{\color[HTML]{333333} Demonstrate for policy change}         & {\color[HTML]{333333} Make optimistic comment}                             \\
{\color[HTML]{333333} Demonstrate for policy change}         & {\color[HTML]{333333} Conduct suicide, car, or other non-military bombing} \\
{\color[HTML]{333333} Obstruct passage, block}               & {\color[HTML]{333333} Appeal for aid}                                      \\
{\color[HTML]{333333} Obstruct passage, block}               & {\color[HTML]{333333} Appeal for humanitarian aid}                         \\
{\color[HTML]{333333} Obstruct passage, block}               & {\color[HTML]{333333} Provide humanitarian aid}                            \\
{\color[HTML]{333333} Acknowledge or claim responsibility}   & {\color[HTML]{333333} Express intent to cooperate militarily}              \\
{\color[HTML]{333333} Acknowledge or claim responsibility}   & {\color[HTML]{333333} Meet at a 'third' location}                          \\
{\color[HTML]{333333} Acknowledge or claim responsibility}   & {\color[HTML]{333333} Demand material cooperation}                         \\
{\color[HTML]{333333} Receive inspectors}                    & {\color[HTML]{333333} Expel or deport individuals}                         \\
{\color[HTML]{333333} Receive inspectors}                    & {\color[HTML]{333333} Express intent to provide material aid}              \\
{\color[HTML]{333333} Receive inspectors}                    & {\color[HTML]{333333} Return, release person(s)}                           \\
{\color[HTML]{333333} Demand release of persons or property} & {\color[HTML]{333333} Use unconventional violence}                         \\
{\color[HTML]{333333} Demand release of persons or property} & {\color[HTML]{333333} Demonstrate or rally}                                \\
{\color[HTML]{333333} Demand release of persons or property} & {\color[HTML]{333333} Appeal for military aid}                             \\
{\color[HTML]{333333} Reject judicial cooperation}           & {\color[HTML]{333333} Appeal to others to settle dispute}                  \\
{\color[HTML]{333333} Reject judicial cooperation}           & {\color[HTML]{333333} Accuse of espionage, treason}                        \\
{\color[HTML]{333333} Reject judicial cooperation}           & {\color[HTML]{333333} Retreat or surrender militarily}                     \\ \bottomrule
\end{tabular}
\caption{Examples of pre-relations and post-relations}
\label{tab:rels}
\end{table*}
\end{document}